\documentclass[twocolumn]{article}
\usepackage{graphicx} % Required for inserting images
\usepackage{booktabs}
\usepackage{hyperref}
\usepackage{authblk}

\title{9th Workshop on Sign Language Translation and Avatar Technologies (SLTAT 2025)}

\author[1]{Fabrizio Nunnari}
\author[2]{Cristina Luna Jiménez}
\author[3]{Rosalee Wolfe}
\author[4]{John C. McDonald}
\author[5]{Michael Filhol}
\author[3]{Eleni Efthimiou}
\author[3]{Evita Fotinea}
\author[6]{Thomas Hanke}

\affil[1]{German Research Center for Artificial Intelligence (DFKI), Saarbr{\"u}cken, Germany}
\affil[2]{University of Augsburg, Augsburg, Germany}
\affil[3]{Institute for Language and Speech Processing, Athena RC, Athens, Greece}
\affil[4]{DePaul University, Chicago, IL, USA}
\affil[5]{Université Paris-Saclay, Paris, France}
\affil[6]{University of Hamburg, Hamburg, Germany}

\date{}

\begin{document}

\maketitle

\begin{figure*}[t]
  \includegraphics[width=\textwidth]{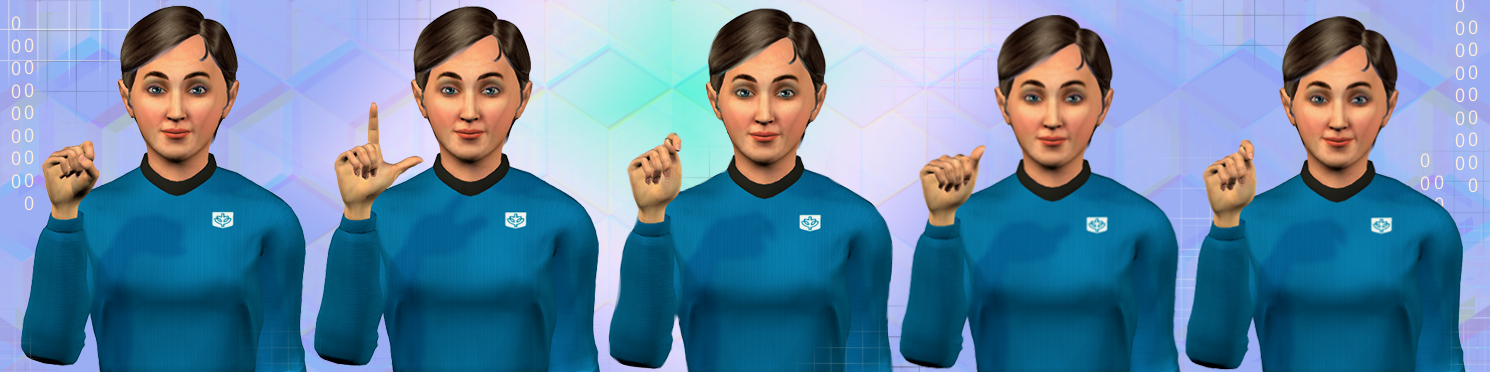}
  \caption{The avatar Paula finger spelling S-L-T-A-T in ASL.}
  \label{fig:teaser}
\end{figure*}

\begin{abstract}
The Sign Language Translation and Avatar Technology (SLTAT) workshops continue a series of gatherings to share recent advances in improving deaf / human communication through non-invasive means.
This 2025 edition, the 9th since its first appearance in 2011, is hosted by the International Conference on Intelligent Virtual Agents (IVA), giving the opportunity for contamination between two research communities, using digital humans as either virtual interpreters or as interactive conversational agents.
As presented in this summary paper, SLTAT sees contributions beyond avatar technologies, with a consistent number of submissions on sign language recognition, and other work on data collection, data analysis, tools, ethics, usability, and affective computing.
\end{abstract}

\section*{Keywords}

Sign language, signing avatars, sign language technology, sign language animation.

\section{Introduction}

Those born deaf constitute an invisible and underserved segment of society \cite{world_helath_organization_deafness_nodate}.  The deaf communities around the world face continual challenges in their daily interaction with hearing, non-signing populations.  The language barrier causes difficulties in accessing health care, education, and job opportunities as well as legal consultation.  For these critical services, the gold standard for facilitating communication has been and still is hiring a sign language interpreter.  However, in daily life, it is impossible for an interpreter to be omnipresent at the many short but important conversations that occur, such as those at a store counter, over a hotel desk, or in an office foyer. An automatic translation system between spoken and signed languages would ease communication obstacles and improve inclusivity while providing a low-cost, non-invasive alternative to cochlear implants \cite{wolfe_sign_2022}.  

The Sign Language Translation and Avatar Technologies workshop, established in 2011, focuses on three main topics: symbolic translation of sign language, animation of sign language using avatars, and usability evaluation of practical translation and animation systems.  At the workshop, a mix of oral presentations as well as poster presentations covering active work and proposed research encourages discussion and collaboration among researchers who come from a wide variety of disciplines, ranging from machine learning and sign language linguistics to mathematics and art.

In a real sense, the SLTAT workshop is coming home this year, because this event was first offered as a standalone workshop in Berlin back in January 2011. Since then, it has been an international symposium with events offered in Germany, Scotland, The United States, France, Canada, and Greece. During this time, it has been hosted at conferences venues such as ACM-ASSETS, ICASSP, LREC, and HCI International. A complete history of the workshop and its past venues can be found on the SLTAT website: \url{http://sltat.cs.depaul.edu}.

This year we are pleased to be welcomed by the 25th ACM International Conference on Intelligent Virtual Agents (IVA2025, \url{https://iva.acm.org/2025/}). This will give the opportunity to get close to a community of researchers who have been focusing on the creation and animation of interactive virtual agents for 25 years. The workshop information for this year can be found on the SLTAT 2025 home page: \url{https://sltat2025.github.io}.

\section{Submissions} 

In total, the workshop achieved 34 submissions of full articles between 4 and 8 pages, of which 30 were accepted and presented at the conference. This year, the workshop achieved a record number of 43 actual reviewers, who reviewed an average of two to three articles after a double-blind peer review process. Among the reviewers, both linguistics and computer science experts participated and enriched the review process.

\begin{figure}
  \centering
  \includegraphics[width=\linewidth]{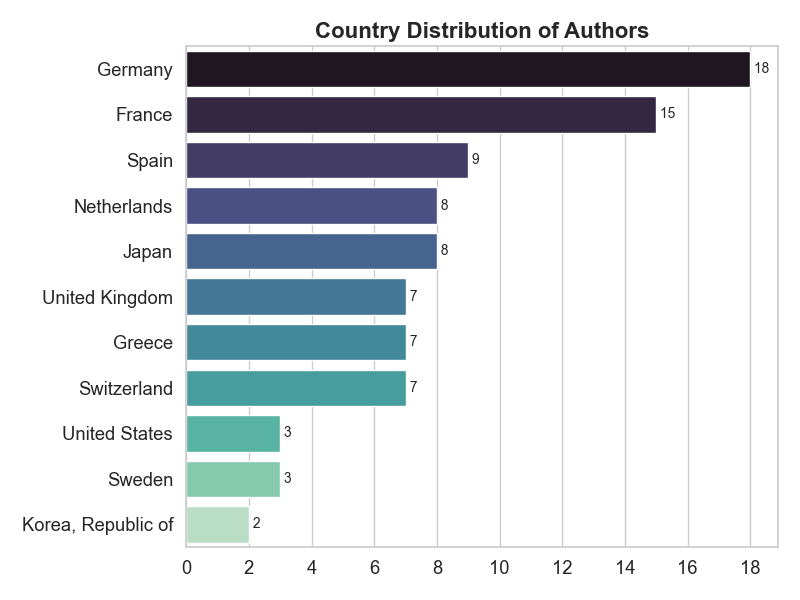}
  \caption{Distribution of authors per country of their institutions}
  \label{fig:authorsCountry}
\end{figure}

A total of 87 different authors participated in the writing and editing process of the 30 submissions, belonging to institutions located in Germany, France, United Kingdom, Japan, Spain, Netherlands, Greece, Switzerland, United States, Sweden, and South Korea; highlighting active Sign Language research communities around the world (see Figure~\ref{fig:authorsCountry}).
% As this year the workshop was hosted in Berlin, we can observe in Figure  a majority of authors sending from institutions in Germany.

The workshop invited submissions in the broad ambit of Sign Language including articles related to translation and recognition technologies, incorporating manual and non-manual features (e.g., mouthing); avatar animation, including linguistically annotation to improve signs animation, as well as flexible facial gestures and mouthing; and in usability, accepting articles focused on evaluating previously mentioned models and editing and preprocessing tools oriented to Sign Language applications. Additionally, this year articles in the ambit of affective computing applied to Sign Language were encouraged, as well as reviews and articles with a focus on ethics and human-centric developments together with the deaf community. The topic distribution is summarized in Table \ref{tab:paper-topics}.

%Topics? How many on avatars? How many on recognition/translation?
%Some trends?

Ten of the 30 accepted articles scoped sign language recognition, translation, or sign spotting. In this regard, the use of models derived from transformers was a prominent resource in the proposed systems; as well as other techniques related to synthetic data as augmentation.

Nine other papers were related to Sign Language Production, avatars, and the evaluation of motion capture systems. 

Five articles belonged to the broad group of datasets, features analysis, and processing tools to automatically edit realistic videos or annotate.

Four articles also addressed relevant ethical issues in sign language and performed a human-centric usability test. Among them, the involvement of the deaf community in research projects and the impact of remote sign language interpretation were discussed.%, and impact of the application and use of SLR tehcnology and its evaluation employing user-experience and acceptability surveys.

Finally, two articles focused on the combined field of sign language and affective computing to address first steps on how to express and combine linguistic features with emotional and facial features to improve sign language prosody.

\begin{table}
    \centering
    \scriptsize
    \begin{tabular}{l|c}
    \toprule
         \textbf{Topic} & \textbf{Count}  \\
         \midrule
         Sign language recognition, translation, sign spotting & 10 \\ 
         Sign language production, avatars, MoCap systems & 9 \\
         Datasets, features analysis, processing tools & 5 \\
         Ethical issues and usability & 4 \\
         Sign language and affective computing & 2 \\
         \midrule
         Total & 30 \\
    \bottomrule
    \end{tabular}
    \caption{Topic distribution of the accepted papers.}
    \label{tab:paper-topics}
\end{table}

% i did not add too much about trends, mainly transformers and LLMs (from the titles) and the use of facial expressions/emotion for prediction and recognition.

\section{Conference organization}

With more than 40 expected participants, the SLTAT workshop has now reached the size of a small conference.
However, rather than switching to a full on-stage conference format, the organizers aim to maximize the chances of face-to-face conversations with other researchers. This, in our opinion, fosters exchange of ideas and collaborations among members of a growing but still tight community.

Thus, the workshop is organized with two short on-stage presentations and two long and populated poster sessions.
This increases the number of work that can be accepted and presented in a 1-day workshop format, leaves space for informal meetings among participants, and last but not least, for the sake of body health, provides a good alternation between sitting and standing positions.

Only six papers will be presented in the two oral/signed on-stage format. The remaining 24 accepted papers will be presented during the two poster sessions.

The workshop organization will provide two International Sign Language interpreters. They will provide interpretation during the on-stage presentation and will be available for 1-to-1 interpretation during the poster sessions.

The schedule of presentations has been devised to take \emph{crip time} into account, addressing the importance of flexible time structures in accessible settings~\cite{ljuslinder_cripping_2020}.

\section{Discussion}

% Avatar tradition
After 40+ years of research \cite{aziz_evolution_2023,naert_survey_2020}, only recently has the use of avatars as the mean of synthesis for sign language started to see application in the industry.
Research on the topic is lively and starts now to focus on improving motion naturalness via better animation techniques and emotional expressivity.

% Tendency towards recognition
In the other direction, the use of neural-based machine learning techniques quickly advanced the recognition of sign language \cite{de_coster_machine_2023,nunez-marcos_survey_2023}.  Research prototypes are already available.

% Better collaboratio with deaf/linguists
However, despite the steadily increasing interest in the topic, there seems to be still an unfair imbalance between the huge number of technologists approaching the topic with respect to linguists and representatives of the Deaf community.
The lack of communication between technologists and linguistically competent researchers has already been recognized and criticized \cite{borstell_ableist_2023,bragg_sign_2019}.

As organizers and coordinators of many initiatives related to SL and technology, we are witnessing increasing awareness (by technologists) of the culture behind signed languages.
However, we believe that future editions will still need to increase the effort for an interdisciplinary work: a better inclusion of the linguistic community for theory-informed technological solutions and a better inclusion of the deaf community to steer technological developments towards the real needs of the every day life of deaf users.

%The goal of this continuing series of workshops is to further progress toward better deaf/hearing communication.  How best to achieve this goal continues to be an open question and contributes to the lively activity in this area of inquiry.

\section{Acknowledgements}
The authors would like to thank the organizers of the Intelligent Virtual Agents (IVA 2025) hosting conference, and in particular the workshop chairs, for all their support: \url{https://iva.acm.org/2025/organizing-committee/}.

This work was partially funded by the German Ministry for Education and Research (BMBF) through the BIGEKO project (grant number 16SV9093).

Avatar montage: Sophia Johnson, Mei Harter.

\bibliographystyle{plain}
\bibliography{SLTAT2025-bibliography}

\end{document}